\documentclass{article}

\usepackage[nonatbib,preprint]{neurips2020_preregistration}

\usepackage[utf8]{inputenc} 
\usepackage[T1]{fontenc}    
\usepackage{hyperref}       
\usepackage{url}            
\usepackage{booktabs}       
\usepackage{amsfonts}       
\usepackage{nicefrac}       
\usepackage{microtype}      

\usepackage{graphicx}
\usepackage{array}

\usepackage{mathptmx} 
\usepackage{amsmath} 
\usepackage{times} 
\usepackage{amsmath} 
\usepackage{amssymb}  
\usepackage{balance}  

\title{Incorporating Rivalry in Reinforcement Learning for a Competitive Game}

 \author{%
  Pablo Barros, Ana Tanevska, Ozge Yalcin, Alessandra Sciutti\thanks{Pablo Barros, Ana Tanevska and Alessandra Sciutti are with the Italian Institute of Technology, Italy. Ozge Yalcin is with the Department of Computer Science, University of British Columbia, Canada.} \\
 }

\begin{document}

\maketitle

\begin{abstract}
   Recent advances in reinforcement learning with social agents have allowed us to achieve human-level performance on some interaction tasks. However, most interactive scenarios do not have as end-goal performance alone; instead, the social impact of these agents when interacting with humans is as important and, in most cases, never explored properly. This preregistration study focuses on providing a novel learning mechanism based on a rivalry social impact. Our scenario explored different reinforcement learning-based agents playing a competitive card game against human players. Based on the concept of competitive rivalry, our analysis aims to investigate if we can change the assessment of these agents from a human perspective.
   


\end{abstract}

\section{Introduction}

The social aspects of interaction are usually dimmed when optimizing an artificial agent through reinforcement learning \cite{liu2019deep}. Most of the training loop must be done in an offline manner \cite{liu2019deep}, or focuses on optimizing objective metrics that do not directly involve social aspects, such as planners \cite{modares2015optimized} or human annotation feedback \cite{churamani2017teaching}. Most of the common success metrics in this regard are related to solving the task in fewer steps, reducing predicted values, or achieving some predefined intermediate objective goals. When the interaction with humans is the main goal, such agents are evaluated mostly based on their objective performance \cite{modares2015optimized}. In the few examples where humans are present in the loop, the success measures are mostly related to the embodied interaction \cite{papaioannou2017combining, gao2019learning}, and not to the underlying decision-making process that these agents learned. 

A great scenario where these problems arise is in competitive interaction. In a competitive game scenario, an agent can learn through reinforcement learning how to adapt towards its opponents \cite{milani2020minerl}, even when these opponents are humans \cite{vinyals2019grandmaster}. However, it is extremely difficult to measure the social aspects of this interaction, without relying on typical human-robot or human-computer interaction schemes \cite{choudhury2019utility}. Although providing important insight on some social aspects, these evaluations usually focus on controlled lab-scenarios \cite{khamassi2018robot}, production of different robotic behavior \cite{ritschel2017real}, and dialogue \cite{cuayahuitl2019ensemble}; this in turn neglects exploring how the agents' various learning strategies influence their explicit behaviour and interaction with humans \cite{tabrez2019improving}, despite it being one of their most important characteristics. This can be evidenced even in the new area of explainable reinforcement learning \cite{madumal2019explainable, sequeira2020interestingness}.


In this preregistration study, we address the problem of including social aspects in the learning strategies of artificial agents in a competitive scenario. We propose an objective human-centered metric, based on rivalry \cite{havard2020rivalry}, to compose the reward function of the agents. Our scenario will be evaluated using the multiplayer Chef`s Hat Card game and will include artificial agents based on Deep Q-Learning (DQL) and Proximal Policy Optimization (PPO). We separate our scenario into two steps: first we try to map the original behavior of such algorithms when playing the game using the novel rivalry score. Second, we predict this metric by having agents identifying human rivals on inference-time, and use this information to update their behavior. To validate the behavior update, and to have a baseline for learning it with the agents, several humans will play the game against the agents and based on a series of explicit and implicit metrics, we will calculate the level of rivalry that each agent yields. 


In this paper, We formalize these problems by proposing the following hypotheses: 
\begin{itemize}

\item \textit{Agents trained with different RL strategies yield distinct rivalry when playing the chef‘s hat card game against humans}
\item \textit{Agents trained with a predicted rivalry as a part of their reward function can modulate human responses on a rivalry scale. }

\end{itemize}

In the remainder of this paper, we detail our motivation, methods, and our experimental setup in a way that it will possible to discuss the possible impact our research has on explainable reinforcement learning and related fields.

\section{Related Work}

\textbf{Reinforcement learning in competitive games}. In the late 1990ies, several researchers tried to identify the impact of the Deep Blue artificial chess player \cite{campbell2002deep} on the development of artificial intelligence \cite{decoste1997future, decoste1998significance}. They all argue that beyond the technical challenge of beating a human, there is an underlying impact on how this agent affects the opponents' behavior during the entire interaction. Over time, these investigations were let aside by the mainstream community, which focused mostly on solving more complex problems. This vision is reflected by the recent development of deep reinforcement learning and the research on training artificial agents to play competitive games that have flourished since \cite{shao2019survey}. AlphaGo \cite{silver2016mastering} demonstrated that these agents can play competitively against humans in very complex games. The recent development of agents that play the StarCraft computer game \cite{vinyals2019grandmaster} pushes these boundaries even further. These agents learn how to adapt to dynamic environments, how to map hypercomplex states and actions, and how to learn new strategies \cite{shao2017cooperative}. Most of the studies, however, focus on the final goal of these agents: how to be competitive against humans. None of them focus on understanding the impact that such agents had on human opponents.

\begin{figure}[t]
    \centering
    \includegraphics[width=0.8\columnwidth]{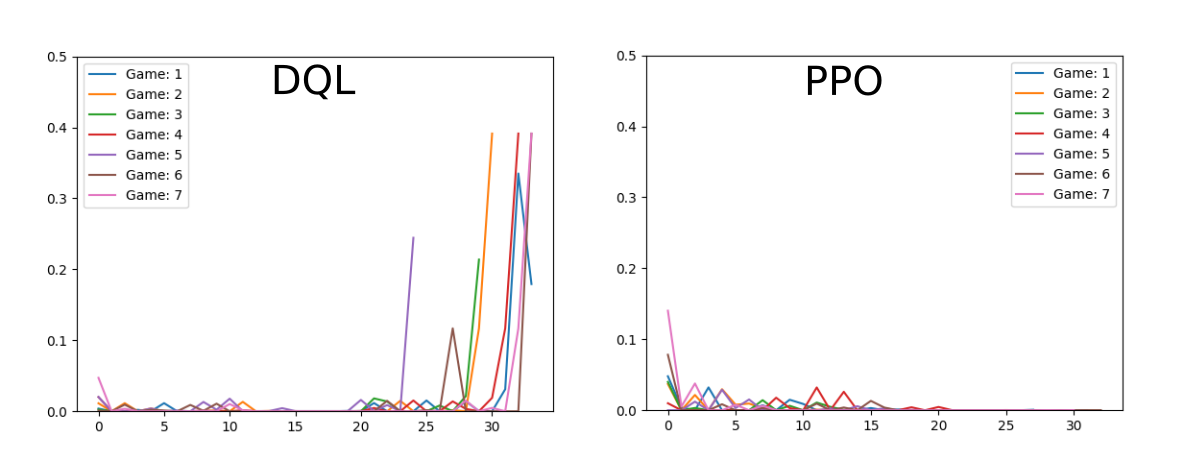}
    \caption{Illustration of the highest Q-Values of an agent trained with DQL and another trained with PPO, playing seven games of the Chef's Hat Card game.}
    \label{fig:dql_ppo_qvalue}
\end{figure}

\textbf{DQL and PPO Playing Behavior on the Chef's Hat Card Game}. In the same development wave, it was recently investigated the design and development of reinforcement learning agents to play the four players Chef's Hat competitive card game \cite{barros2020ad}. These agents were based on Deep Q-Learning (DQL) \cite{van2016deep} and Proximal Policy Optimization (PPO) \cite{schulman2017proximal}, and achieved success in learning how to win the game in different tasks: playing against random agents, self-play, and online adaptation towards the opponents. It was observed, however, that these agents present different behavior during game-play while maintaining a similar objective performance measured by overall wins over a series of games. By observing the Q-values of each agent during an entire match, it is possible to see in Figure \ref{fig:dql_ppo_qvalue} that the DQL agent usually presents higher Q-values at the end of a game, while the PPO agent presents them at the beginning of the game, but with lower intensity. This led to a general understanding that, although both effective, these agents learned different strategies. What has not yet been done is to measure the social impact that such strategies can have when the agents play against humans.

\textbf{Human-centric Analysis of RL}. When analyzing the impact of artificial agents on humans, there are now decades of studies focused on Human-Robot \cite{rossi2017user} and Human-Computer \cite{arzate2020survey} Interaction (HRI and HCI respectively). Most of these studies, however, focus on optimizing RL agents to solve a specific task, even social ones, without having much feedback on the social aspects of the task as part of their learning mechanism. Such agents are usually designed to learn an expected outcome, such as improving engagement \cite{papaioannou2017combining}, or to imitate humans \cite{milani2020minerl}. None of the most recent studies focus on extracting the intrinsic behavior bias that different learning schemes apply to the final agent.

\textbf{Rivalry in Competitive Games}. One way to explain the behavior of an agent as a factor of its learning strategy is to measure its impact on humans. There exist several social metrics that take into consideration the interaction \cite{murphy2013survey}, but most of them focus on the subjective impressions humans have on the embodied interaction \cite{canamero2020embodied}, the subjective quality of the interaction \cite{peltason2012talking} and/or the efficiency of the interaction to solve a task \cite{begum2015measuring}. In a competitive game, however, one of the most informative metrics is the rivalry \cite{havard2020rivalry} between the human and the agents. It informs us if the human recognizes the agent as a threat, as an engaged interaction partner, and as a same-level competitor \cite{furrer2000rivalry}. These characteristics involve many different levels of perception and world-understanding but can be measured directly from the way that artificial agents play a game if we exclude any type of embodied perception and language-based communication, for example. In this regard, utilizing a human-centric metrics like the rivalry to assess the learning strategies of an artificial agent is a novel perspective that can bring in a more intuitive understanding of the agents' behavior.


In competitive games, the rivalry is a central concept that directly effects the opponent's behavior through their motivation in play. In human-to-human scenarios and economics, healthy rivalry is considered to be an important factor that can positively affect the performance of opponents; while in other situations it can also contribute to unnecessary risk-taking behavior. 
In human-in-the-loop online learning scenarios for competitive games, the absence of rivalry or competitiveness might result in the human opponent to lose motivation to play the game and show sub-optimal performance during game-play. The agents who are learning actively from the human, are bound to learn from this bad performance input which in turn would result in sub-optimal learning. By introducing the notion of rivalry, we aim to evaluate how different agents affect the user perception of the agents and in turn have an effect on user performance based on the increased competitiveness and rivalry effects.

\section{Methodology and Experimental Protocol}

\textbf{The Chef's Hat Card Game  \cite{barros2020food} } introduces a competitive multi-player scenario designed to be used both in human-human and human-robot interaction (HRI).
As a reinforcement learning task, it provides a controllable action-perception cycle, where each player can only perform a restricted set of actions. This in turn allows each player to behave as organically as possible and allows the direct measure transfer from a real-world game into an OpenAI GYM-ready scenario without any functionality loss. 

Chef's Hat is a 4-player round-based card game, where each person has a restaurant-context role (Chef, Sous-Chef, Waiter or Dishwasher). This role is updated after each game based on the order of finishing the previous match. At the beginning of the game, the players are dealt a full hand of cards (17 cards per player), and taking turns they need to dispose of their cards as quickly as possible. The cards represent ingredients for pizzas, and each round consists of the players making pizzas by discarding cards in a certain manner. Wins the match the first player that discards all its cards, receiving 3 points. The second player receives 2 points, the third receives 1 point and the last player receives no points. Wins the game the first player to reach 15 points. The details of the pizza-making rules and the role hierarchy are explained fully in the games' formal description \cite{barros2020food}.

\textbf{Reinforcement Learning Agents}. To simulate players in the game, we must train reinforcement learning agents. Following the original design proposed by the Chef's Hat authors \cite{barros2020ad}, we will implement and train two types of agents: one based on Deep Q-Learning (DQL) and another one based on Proximal Policy Optimization (PPO). We need to implement these agents as their learned behavior will be our baseline for our final evaluation.

The Chef's Hat simulation environment defines the state space as an aggregation of the cards on the board and the current cards in the player's hand. The action space is composed of 199 allowed discard actions (which comprise each card and card combinations allowed by the game rules), and one pass action, totaling 200 actions. Each agent is trained using the same reward: 1 for an action that leads to a victory, and -0.01 for every other action. Using such ultimate reward guarantees that the agents will have to explore enough and find their strategies to win the game.

We will train the agents from scratch, first playing against agents that output random movements, and later on against themselves. As each game is composed of 4 players, we will mix different instances of each agent in different game sessions. The entire optimization of the agents, including topology definition, training parameters, and training strategies will be defined using standard optimization schemes, such as grid-search and TPE. As our goal is to obtain agents that learn how to win the game, the entire optimization for the first experimental task will be done based on a pure objective performance: the total number of victories in a set of games.

\textbf{Data Collection and Experimental Environment}. Our experimental environment involves the agents playing the game against a human opponent. This will be done with a web-based interface for the Chef's Hat simulation environment, that will allow a human to play, in real-time, against three other artificial agents. The web-interface will be bounded to the same rules as the game's simulation, to maintain the same evaluation environment. Each human will play a full 15 points games against three agents, one based on DQL, one based on PPO, and one based on an agent that chooses random moves. For each match, we will collect all the objective and state information from all players, but most importantly the total game points ($points$) per match. We will also collect subjective information from the human players, to understand better their behavior: 

\begin{itemize}
    \item Before the game starts, we will proceed with a simple questionnaire to self-assess the person's agency ($ag_h$), competence ($ct_h$), sense of communion ($cm_h$) \cite{Eagly_Nater_Miller_Kaufmann_Sczesny_2020} and competitiveness ($C_h$) \cite{smither1992nature} in a scale between 0 and 5. 
    \item At the end of the game, the human player will assess each agent adversary based on their agency ($ag_a$), competence ($co_a$), and communion ($cm_a$) traits. 
\end{itemize}

Agency, competence, and communion were chosen as representative of personality traits due to their simple understanding, which facilitates the own and third-party assessment \cite{abele2007agency}. Also, due to its direct mapping between action and perception \cite{carrier2014primacy}, which makes it ideal for a competitive card game analysis.

\begin{figure}
\centering
\includegraphics[width=0.5\columnwidth]{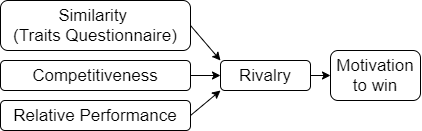}
\caption{The theoretical model of rivalry used for our new rivalry metric proposition, based on the framework proposed by Kilduff et al. \cite{kilduff2010psychology}.}
\label{fig:rivalry}
\end{figure}

    
\textbf{Proposing Rivalry}. To optimally define the impact that each agent has on the players, we will use a novel formalization of Rivalry \cite{kilduff2010psychology}. Rivalry can be defined as a subjective social relationship arising between two actors based on the competitive characteristics of an individual, as well as the increasing stakes and psychological involvement in the situation. Thus, a proposed theoretical model of rivalry, illustrated in Figure \ref{fig:rivalry}, suggests that antecedents of rivalry are similarity factors, competitiveness, and relative performance of the agents \cite{kilduff2010psychology}. In our scenario, this translates into:

\begin{equation}
S_a =  \sqrt{(ag_h-ag_a)^{2} + (ct_h-ct_a)^{2} + (cm_h-cm_a)^{2}}
\end{equation}

\begin{equation}
P_a =  (points_h - points_a) / 15
\end{equation}

\begin{equation}
R_a =  (S_a + C_h + P_a) / 3
\end{equation}

\noindent where the index $a$ defines one of the agents, the index $h$ defines the human ratings, $S_a$ is the similarity between the player and the agent, assessed by the player themselves; $P_a$ is the relative performance for each agent, and $R_a$ is the final rivalry rating. 

The presence of a rivalry effect affects the motivation of the individual and their performance. Thus, we expect that rivalry rates will change accordingly to the game development. As the users will evaluate the entire game behavior of an agent, and the only difference between players is the way they play the game, the measure of rivalry will reflect directly the user's perception of the outcome of the reinforcement learning strategies. 

\textbf{Predicting Rivalry}. In our second experiment, each agent will calculate rivalry against the human player. To that, they follow the same rivalry calculation as humans. To predict similarity, tho, the agents will use a similarity predictor that will be trained based on the humans´ responses. The predictor ($pr_a(h)$) will match state+action chosen by humans ($h$) with the given personality traits ($ag_a$, $ct_a$ and $cm_a$), obtained previously in the data collection. The similarity predictor of each agent will infer, during game-play, the personality traits of each human the agents play against. Each type of agent, DQL, and PPO will also have a single set of personality traits associated, given also by our human analysis of scenario one. Thus the similarity based on an agent´s ($a$) perspective ($S_h$) is given as:

\begin{equation}
S_h = \left ( \sqrt{(pr_a(h) - (ag_a, ct_a, cm_a) )^{2}} \right )
\end{equation}

The performance measure of the agents will be given by their own assessment of their actions. To achieve this, each agent will compute the introspective confidence ($ic_a$) \cite{cruz2020explainable} of each action, which focuses on scaling the selected Q-value of an action towards the final goal using a logarithm transformation which computes the probability of success, in our case, of winning the game. The introspective confidence gives us a self-assessment of the agent's actions, based on its own game experience. The agent´s perspective is given as:

\begin{equation}
C_a = \frac{\sum ic_a (act) }{totalActions}
\end{equation}

\noindent where $act$, is each of the actions the agent took during the game.

The relative performance ($P_h$) is calculated similarly to the human´s perspective, but taking the agent´s perspective into consideration. Thus, the predicted rivalry ($R_h$) is defined as the mean of $(S_h + C_a + P_h)$.

To include the predicted rivalry in the reward function will impact on the agent behavior. As the rivalry prediction is a representation of subjective and objective human metrics,

\textbf{First Experimental Scenario}. Our first experimental scenario involves one human playing the game against three agents: two pre-trained agents, one using DQL and the other PPO, and an agent that only does random moves. Our goal with this experiment is to collect the human assessments about the agents, to provide an appropriate dataset for the rivalry prediction. Also, it will be important to identify the baseline assessment that each of the trained agents already carries, which will be completely related to their learning strategy during training.

\textbf{Second Experimental Scenario}. We will repeat the same setup as the first experimental scenario, except the following: each agent will be updated, during game-play, based on the novel reward function that includes a rivalry score. The rivalry score will be predicted by each agent in a process defined above. Each agent will include the rivalry against the human in its reward function, to change the human perception toward themselves. The agents will be trained once per finished match. For each agent, we will experiment with three conditions: increase rivalry, decrease rivalry, and maintain a rivalry.

\textbf{Expected Contribution} We believe that integrating the social representations given by our novel rivalry score will enrich current reinforcement learning strategies. We hope to pave the way to bridge the gap between pure reinforcement learning optimization and social learning applications.

\bibliographystyle{unsrt}
\bibliography{sample-base}

\end{document}